\documentclass{article}
\usepackage{ijcai20}

\usepackage{times}

\usepackage{soul}
\usepackage{url}
\usepackage[hidelinks]{hyperref}
\usepackage[utf8]{inputenc}
\usepackage[small]{caption}
\usepackage{graphicx}
\usepackage{amsfonts}
\usepackage{amsmath}
\usepackage{booktabs}
\usepackage{multirow}
\usepackage{xcolor}
\usepackage[linesnumbered,ruled]{algorithm2e}
\title{Meta Segmentation Network for Ultra-Resolution Medical Images}

\author{
	Tong Wu$^1$\footnote{Equal contribution.}\and
	Yuan Xie$^2$\footnotemark[1]\and
	Yanyun Qu$^1$\footnote{Corresponding author.}\and
	Bicheng Dai$^1$\footnotemark[2]\and
	Shuxin Chen$^1$\\
	\affiliations
	$^1$Fujian Key Laboratory of Sensing and Computing for Smart City, School of Informatics\\
	$^2$School of Computer Science and Technology, East China Normal University, Shanghai, China\\
	\emails
   	tongwu@stu.xmu.edu.cn,   
   	yxie@cs.ecnu.edu.cn,
   	yyqu@xmu.edu.cn,
   	nejordai@163.com,
   	chenshuxin@stu.xmu.edu.cn      
}

\begin{document}

\maketitle

\begin{abstract}	
Despite recent progress on semantic segmentation, there still exist huge challenges in medical ultra-resolution image segmentation. The methods based on multi-branch structure can make a good balance between computational burdens and segmentation accuracy. However, the fusion structure in these methods require to be designed elaborately to achieve desirable result, which leads to model redundancy. In this paper, we propose Meta Segmentation Network (MSN) to solve this challenging problem. With the help of meta-learning, the fusion module of MSN is quite simple but effective. MSN can fast generate the weights of fusion layers through a simple meta-learner, requiring only a few training samples and epochs to converge. In addition, to avoid learning all branches from scratch, we further introduce a particular weight sharing mechanism to realize a fast knowledge adaptation and share the weights among multiple branches, resulting in the performance improvement and significant parameters reduction. The experimental results on two challenging ultra-resolution medical datasets BACH and ISIC show that MSN achieves the best performance compared with the state-of-the-art methods.

 
\end{abstract}

\section{Introduction}
With the rising up of deep learning, semantic segmentation achieves prominent progress. However, the semantic segmentation of ultra-resolution image (URI) is seldom studied, especially in the application of medical diagnosis. Many medical URIs \cite{ISIC-1,BACH} contain more than $4$ M pixels per image, and as for whole-slide image (WSI), a special type of medical URIs, its size even exceeds $40000\times60000$ (about $30$ M pixels). The URIs with huge size require large computational resources, which some most popular semantic segmentation framworks, such as UNet \cite{UNet}, PSPNet \cite{PSPNet}, and DeepLab \cite{DeepLab-V1,DeepLab-V3+}, are hard to afford.


There are two common ways to process URIs: image downsampling and sliding patches \cite{patch_method1,patch_method2}. The former resizes a large image to a suitable size, {\it e.g.,} $512\times512$, then feeds it into the model, which leads to the great loss of local details, especially for WSIs. The latter crops original image into many small patches, then segments on patch-level, and finally combines the segmentation results of these patches. While these methods can effectively reduce the computational burden, the global information that provided by spatial context and neighborhood dependency is almost abandoned, which makes it difficult to obtain accurate segmentation results.
\begin{figure}[]
	\center{\includegraphics[width=.95\columnwidth]{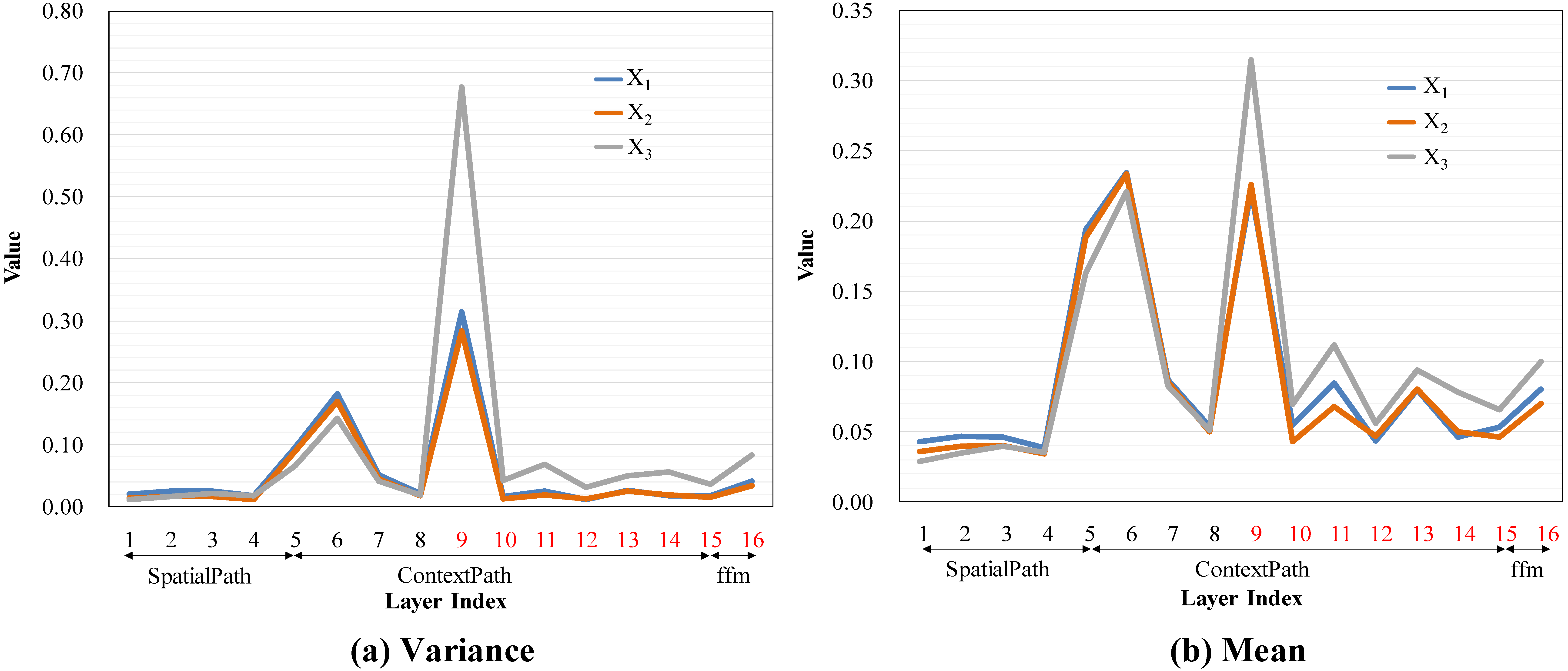}}
	\caption{The statistics of mean and variance of the convolutional activations for the backbone network (BiSeNet). We train the branch of $X_3$, and fix the parameters. Then we sequentially input $X_1$, $X_2$ and $X_3$ to calculate their mean and variance. The layer index in red font indicates the `gap layer'.}
	\label{mean_variance_bise}
\end{figure}

The latest representative patch-based method is the AWMF-CNN \cite{AWMF-CNN}, it is a multi-branch structure to aggregate contextual information from multiple magnification patches which contain target regions and receptive fields in different resolutions and scales.
\emph{As a popular strategy, multi-branch induced methods find a tradeoff between small inputs and multi-scales, however, they inevitably introduce two challenge problems}: Firstly, they usually need a carefully designed fusion mechanism for final result, {\it e.g.,} the fusion layers consist of many stacked convolutions with relatively more channels or the auxiliary weighting net in \cite{AWMF-CNN}, resulting in a complicated and redundant structure. Fortunately, \emph{with developing a fusion method via meta-learning}, only a simple structure is needed in our method to ensure good results. Secondly, all the branches are independent so as to be trained from scratch separately, increasing overall parameters significantly.

In this paper, we propose a novel multi-branch based framework guided by a meta-learning way for ultra-resolution medical images segmentation, namely Meta Segmentation Network (MSN). Recently, meta-learning has attracted increasing attentions. The negative loss gradient, which contains more detail information, {\it e.g.,} the target-specific difference between prediction and label, can be used as a very useful information for fast weights generation for convolution, as it has been confirmed in \cite{Meta-learning1}. Moreover, the structures of most meta-learning frameworks are quite simple but highly effective \cite{MAML,MetaPruning},  hence the elaborative structural design is unecessary. 

For this purpose, we develop a meta-fusion mechanism, which can elegantly solve the first challenge in multi-branch methods. Specifically, we use the negative gradients of the output layers of branches as the meta-information to train a meta-learner, and directly predict the weights of the fusion layer.  Our method is superior to the training way of traditional end-to-end BP that needs more iterative steps and more training samples to converge. It is also noticeable that, unlike the elaborately designed fusion structure in AWMF-CNN, {\it i.e.,} many stacked convolution layers as well as a redundant weighting net, the structure of meta-fusion contains only two convolutions with a few channels and a simple meta-learner.

To avoid learning all the branches from scratch, we further introduce a quite effective weight sharing mechanism. Although the inputs for those branches have different magnifications, they are still in the same domain. So, we believe that the knowledge among these branches can be shared to some extent, {\it i.e.,} weight sharing. In our weight sharing, we adopt a special memory mechanism to achieve fast knowledge adaptation between meta-branch and non-meta-branches. The meta-branch represents a reference branch containing basic parameters that need to be shared with other branches. Moreover, we experimentally find that direct weight sharing leads to some knowledge gaps, as illustrated in Fig. \ref{mean_variance_bise}. To bridge these gaps, we use the memory mechanism to store some useful memory (feature) from the meta-branch, then make a memory transformation between meta-branch and non-meata-branches to realize a fast knowledge adaption.

The contributions can be summarized as follows:
\begin{itemize}	
	\item Meta-fusion mechanism is proposed for a multi-branch deep model for URI segmentation by utilizing meta-learning. The weights of fusion layer can be fast generated through the meta-learner, leading to a simple but highly effective model.
	\item  A novel weight sharing mechanism is introduced to realize fast knowledge adaptation, resulting in significant reduction in training process and overall parameters.
	\item The proposed MSN achieves the best performance on two challenging datasets: BACH and ISIC. Especially, our method achieves a significant performance improvement over the latest AWMF-CNN, and the overall parameters are close to that of a single branch. Thus, it is a practical segmentation method both in resource-saving and accuracy.
\end{itemize}

\begin{figure*}[!t]
	\center{\includegraphics[width=1.95\columnwidth]{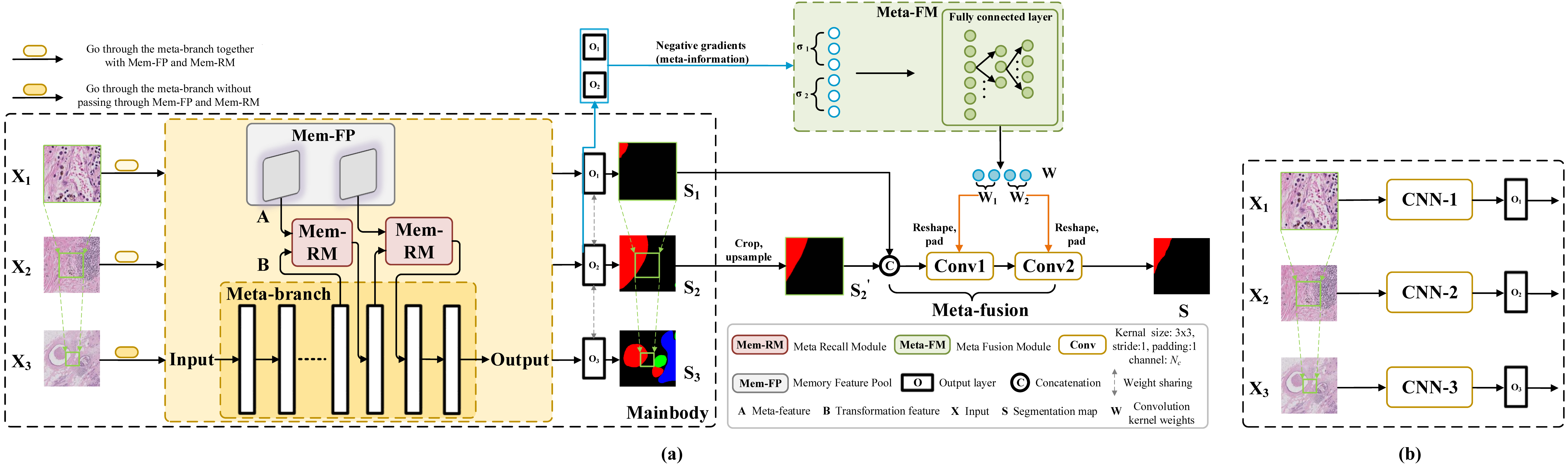}}
	\caption{ (a) The architecture of MSN. (b) The expanded structure of Mainbody in (a) without weight sharing, where three branches are separated. MSN mainly contains two components: Mainbody and Meta-FM, where Mainbody contains three key components: the meta-branch, Mem-FP and Mem-RM. Mainbody receives different resolution image patches as input, and outputs their segmentation maps. The input $X_3$ only go through the meta-branch without passing through Mem-FP and Mem-RM, while the other two resolution inputs go through the meta-branch together with the integrated Mem-FP and Mem-RM to fix the gap layers.  Meta-FM is to fuse the results of the branches in a meta-learning way, where the output channels of two fusion convolutions are both $N_c$, and $N_c$ is the number of classes.}
	\label{MSN}
\end{figure*}
\begin{figure}[!t]
	\center{\includegraphics[width=0.80\columnwidth]{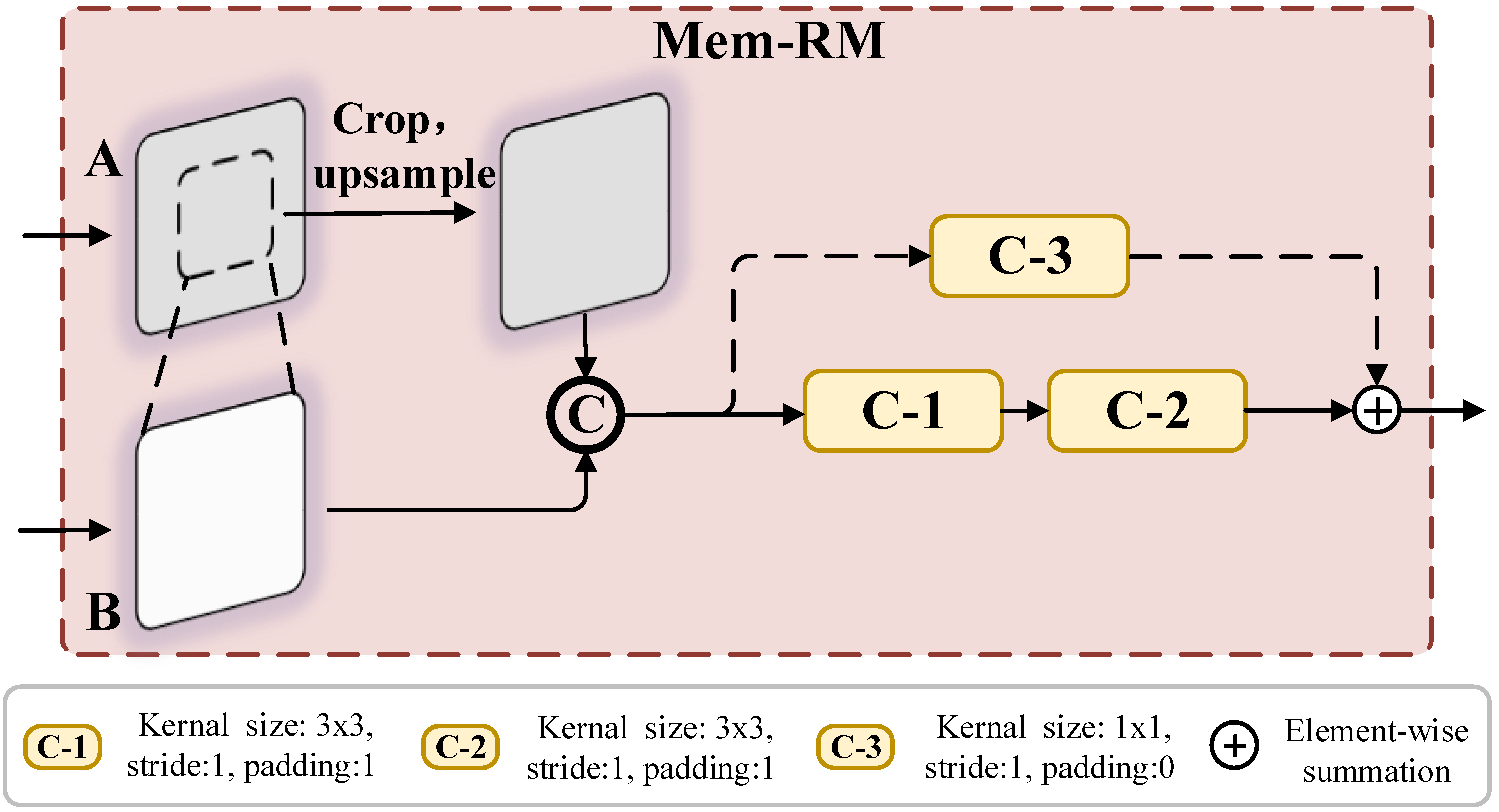}}
	\caption{The architecture of Mem-RM. Mem-RM is added to the gap layers in the meta-branch. Mem-RM utilizes the meta-features which are stored in Mem-FP to realize the ``memory recall''.}
	\label{Mem-RM}
\end{figure}

\section{Proposed Method}
\subsection{Architecture of MSN} \label{architecture of MSN}
In this section, we introduce the framework of MSN, whose architecture is illustrated in Fig. \ref{MSN}. It mainly contains two components: the multi-branch structure, named Mainbody and the Meta Fusion Module (Meta-FM).
Mainbody is an all-in-one structure which realizes multi-resolution segments. Unlike the general multi-resolution structure which requires the multiple branches with a special resolution per branch, Mainbody only uses one branch to realize the multi-resolution segmentation. Let's name the branches as the high-resolution, middle-resolution and low-resolution branches corresponding to the counterpart resolution inputs. Mainbody containts three key parts: the meta-branch, the Memory Feature Pool (Mem-FP) and the Memory Recall Module (Mem-RM).  Mem-FP stores the meta-features of the meta-branch in Mainbody, while Mem-RM is deployed in non-meta-branch to complement the distinctive features from the non-meta branch with the meta-features stored in Mem-FP, named Memory.

Mainbody outputs two preliminary segmentation maps, {\it i.e.,} $S_1$ and $S_2$ in Fig. \ref{MSN}, which will be fused in a meta-learning way to achieve a final segmentation result. In the following, we will detail the operating mechanism of Mainbody, Mem-FP, Mem-RM, and Meta-FM.

\paragraph{Mainbody.} 
The architecture of Mainbody is shown in the middle part of Fig. \ref{MSN}.
Let $X_1$, $X_2$ and $X_3$, denote the three types of inputs with the same size having different resolution corresponding to the three branches ({\it e.g.,} $16\times$, $4\times$, $1\times$). Actually, $X_3$ has the widest receptive field with the lowest resolution,  while $X_1$ is the opposite. As shown in Fig. \ref{MSN}, $X_1$ is the upscaled patch centered in $X_3$ signed in a green box and has the high-resolution and $X_2$ has the middle-resolution.  Mainbody can process the image patches with three resolutions and output the counterpart segmentation maps $\{S_1, S_2, S_3\}$.  

Considering the commonness and difference among the  knowledge of the multi-resolution segmentations, we treat the low-resolution branch as the meta-branch which share the weights with the middle-resolution and high-resolution branches because it contains the most information, and use Mem-FP and Mem-RM to adjust the weight learning. In detail, a low-resolution image patch $X_3$ is fed into Mainbody, and passed through the non-gap convolution layers and gap convolution layers in the meta-branch. In gap layers, the obtained feature maps are recorded in Mem-FP. As for the high-resolution image patch $X_1$, after fixing all the layers of the meta-branch, it is passed through Mainbody just like $X_3$ in non-gap convolution layer, when meeting gap convolution layer, Memory is recalled from Mem-FP, and Memory as well as the feature maps output by the current gap layer are fed into the Mem-RM for adjusting the weight learning. So do the middle-resolution image patch $X_2$. Subsequently, we fuse the two outputs $S_1$, $S_2$ of non-meta-branches in Meta-FM. In the following, we introduce the Mem-FP, Mem-RM,and Meta-FM. 

\paragraph{Memory Feature Pool.} Mem-FP acts as a storage pool. As shown in Fig. \ref{mean_variance_bise}, the branches of $X_1$ and $X_2$ have a large gap with $X_3$ (meta-branch) in some layers of CNN. When processing the low-resolution image patch $X_3$, feature maps output by the gap layers in the meta-branch, named as Meta-features, is saved in Mem-FP which will be utilized to compensate other branches. Actually, once $X_3$ is passed through 
the meta branch, the obtained Meta-features are stored.

\paragraph{Memory Recall Module.} In order to make the meta-branch adapt to other branches in the weight sharing mechanism, non-meta-branches should ``recall'' the missing features in Mem-FP at these big gap layers. Therefore, we construct Mem-RM and embed it in the meta-branch as an auxiliary module to recall the memory of the gap layers in the meta-branch. Specifically, as shown in Fig. \ref{Mem-RM}, when an image patch is fed into the non-meta-branch, such as $X_1$ or $X_2$, it is passed forward along the meta-branch until it meets the gap layers. At the gap layers, both the pre-saved meta-feature and the counterpart feature from non-meta-branch are fed into Mem-RM. As shown in Fig.3, there are two input branches: the top branch inputs Meta-feature of $X_3$ (A) and the bottom branch inputs the output feature maps of $X_1$ or $X_2$ (B). In order to align the feature maps between A and B, we crop the target region centered in Meta-feature and upscaled to the same size as B. After that we concatenate them and implement  convolutions on them. The process is formulated as:
\begin{equation}\label{Mem-RM}
\hat{B} = f(cat(B, up(crop(A)))), 
\end{equation}
where $\hat{B}$ is final output of Mem-RM, $f(\cdot)$ is the nonlinear transformation function, and $cat(\cdot, \cdot)$, $up(\cdot)$ and $crop(\cdot)$ are the operations of concatenation, upsampling and cropping, respectively.

\paragraph{Meta Fusion Module.}
The final step of our framework is the fusion of different branches. Since the branches of $X_1$ and $X_2$ have already captured the memory of $X_3$, we only need to consider the fusion of the branches of $X_1$ and $X_2$. One of the most common way is to use an elaboratively designed structure that might contain dozens of convolution layers to perform feature fusion, and then conduct common optimizatizer, {\it e.g.,} SGD \cite{SGD}, to adjust the parameters of these convolutional layers. This process may take many iterations for optimization to achieve convergence. 

To pursue a compact and simple but highly effective fusion structure, we propose using a specific target provided by an auxiliary meta-learner. It is well known that, the negative gradient, which is used in SGD to determine the direction of optimization, contains the detail information that measure the difference between prediction and ground truth. Draw lesson from the theory of negative gradient, we construct Meta-FM to predict the weights of these convolutional layers directly. The structure of Meta-FM is shown in Fig. \ref{MSN}, and Meta-FM receives the negative gradients (meta-information) of the output layers of two branches, and output the predicted weights through two fully connected layers (FC). Meta-FM can be formulated as: 
\begin{equation}\label{Weight generation}
W = [W_1, W_2] = f(\sigma), 
\end{equation}
where $W_1$ and $W_2$ are the parameters of two fusion convolutions, respectively. Note that, $W_1$ and $W_2$ should be reshaped to the weight matrixs because the output of FC is a vector. $f(\cdot)$ is a nonlinear function which contains the structure of FC-Relu-FC. $\sigma$ is the gradient vector of the output layers of two branches, which is formulated as:
\begin{equation}\label{Meta-information}
\sigma = cat(v(-\frac{\partial L(S_{1},Y_1)}{\partial W_{o1}}), v(-\frac{\partial L(S_{2}^{'},Y_1)}{\partial W_{o2}})), 
\end{equation}
where $L$ is the loss function, $S_{1}$ is the segmentation prediction of the branch of $X_1$, and $Y_1$ is the ground truth. $S_{2}^{'} = up(crop(S_2))$, since the resolution of $X_2$ is lower than $X_1$, we crop the target region from $S_2$ and upsample it to the size of $S_1$. $W_{o1}$ and $W_{o2}$ are the weights of the output layers of the branch of $X_1$ and $X_2$, respectively. The operation $v(\cdot)$ reshapes the gradient matrix to a column vector.

\subsection{Loss Function}
We use the cross entropy as the loss function of our model, which can be formulated as:
\begin{equation}\label{loss function}
L(P, Y) = -\sum_{i}^{N}\sum_{j\in P_{i}}Y_{i, j}\log P_{i, j}, 
\end{equation}
where $P$ is the predicted segmentation maps, $Y$ denotes the counterpart ground truths, $N$ is the total number of samples, and $j$ is the $j$-th pixel of $P_i$. This loss funtion will be used in multiple segmentation results of our model, {\it i.e.,} $S_1, S_2, S_3$ as well as the final fused result $S${\color{red},} to train our model.

\subsection{Training}
We adopt a $3$-step training scheme to train MSN. Step $1$. we train the meta-branch in Mainbody to obtain the meta parameters that will be shared with other branches. Step $2$. the Mem-RM is trained for the non-meta-branches to fix knowledge gaps. Step $3$. Meta-FM is learned to fuse the multi-resolution segmentation results. We divide the training data into two parts: training set and sub-training set. The sub-training set is much smaller than the training set. The training set is used for the first step, while the sub-training set is involved in the second and third steps. We initialize all layers similar to \cite{kaiming}.

\paragraph{Training Meta-branch.}  The low-resolution image patch $X_3$ is fed into the branch and obtain the segmentation map $S_3$, and then the weights of this branch is updated with the loss function $L(S_3, Y_3)$ formulated in Eq. (\ref{loss function}).
\paragraph{Training Mem-RM.} After Step $1$, we have the meta parameters and fix them. Next we train Mem-RM on sub-training set to alleviate the influence of gap layers w.r.t the meta-branch. We input $X_1$ or $X_2$ to the fixed meta-branch with their specific Mem-RMs and get the counterpart segmentation results $S_1$ and $S_2$. Then we use the loss function $L(S_1, Y_1)$ and $L(S_2, Y_2)$ to update each Mem-RM which is specific to the branch of $X_1$ or $X_2$. 
\paragraph{Training Meta-FM.} Firstly, we fix the trained meta-branch and Mem-RM. Then we obtain the segmentation maps $S_1$, $S_2$ by using the same way of Step $2$. After some operation as mentioned before, {\it e.g.,} cropping and concatenation, finally we feed the processed $S_1$ and $S_2$ to the fusion layers whose weights are generated by Meta-FM, and obtain the fusion result $S$. Since the reshape operation on the weights vector before padding into fusion layer is differentiable, we thus tune the parameters of Meta-FM in few epochs by minimizing the loss function $L(S, Y_1)$ on sub-training set. 
\section{Experiments}
In this section, we evaluate our method on two ultra-resolution medical datasets: BACH and ISIC.  We take two criteria for evaluation: the mean Intersection over Union (mIoU) and the amount of model parameters.
\subsection{Datasets}
\textbf{BACH} \cite{BACH} is composed of Hematoxylin and Eosin (H$\&$E) stained breast histology microscopy and whole-slide images (WSI). There are $10$ WSIs, with an average size of $42113\times62625$ pixels (about $3000$ M pixels), included in BACH.  These WSIs are stored in a multi-resolution pyramid structure, {\it i.e., } $1\times$, $4\times$ and $16\times$. Four classes are presented in BACH: normal, benign, in situ, and invasive carcinoma. We randomly split $10$ WSIs into $7$, $1$, $2$ images for the training set, the sub-training set and the test set, respectively. 

\textbf{ISIC} \cite{ISIC-1,ISIC-2} is an ultra-resolution medical dataset for pigmented skin lesions, which total contains $2596$ images. Its average resolution is up to $9$ M, while the highest resolution is up to $6748\times4499$. The dense annotations contain two classes: lesion, normal. We randomly divide the dataset into training, sub-training and testing sets with $2077$, $360$ and $157$ images.
\subsection{Implementation Details}
In our model, we use BiSeNet \cite{BiSeNet} as backbone, {\it i.e.,} the CNN structure in Mainbody which contains three branches: the high-resolution branch, the middle-resolution branch and the low-resolution branch. We feed the patch with the size of $256\times 256$ into MSN. We firstly crop out $X_1$ from left to right in image without overlapping except the last patch in each row. Then we align the center of the  target area to crop out $X_2$ and $X_3$, if the cropping patch exceeds the boundary, then $0$ is padded. As for BACH, we use the professional tool ``OpenSlide'' \cite{OpenSlide} to read the multi-resolution pyramid in WSI, where the resolutions of the input patches fed into the three branches are $16\times$, $4\times$, and $1\times$, respectively. We finally get training, sub-training and test set with $9520$, $2379$ and $75603$ patches for each resolution. As for ISIC, we set three resolutions as $4\times$, $2\times$ and $1\times$, where the original image is considered as the highest resolution. Then we crop out $X_1$, $X_2$ and $X_3$ in these three-resolution images, respectively. The number of patches in each resolution for training, sub-training and test set are $21471$, $3001$ and $52166$, respectively. We train the meta-branch for $30$ epochs, and tune the non-meta-branches as well as Meta-FM for only $10$ epochs, with the batch size of $32$. The optimizer Adam \cite{Adam} is utilized with initial learning rate $0.0001$ to update the parameters of network. The whole model is trained in PyTorch \cite{Pytorch} with a single $1080$Ti GPU.

\begin{table}[!t]
	\centering
	\resizebox{.95\columnwidth}!{\begin{tabular}{@{}lrrrrr@{}}
			\toprule
			\multirow{2}{*}{Method} & \multicolumn{4}{c}{mIoU (\%)} & \multirow{2}{*}{$\#$ Parm (M)} \\ \cmidrule(lr){2-5}
			& $X_1$    & $X_2$    & $X_3$   & Fusion &                                 \\ \midrule
			UNet (2015)             & 25.2  & 31.7  & 37.9 & -      & \textbf{7.8}                             \\
			PSPNet (2017)           & 29.0  & 36.3  & \textbf{45.3} & -      & 48.8                            \\
			DeepLab-V3+ (2018)      & 26.6  & 30.8  & 42.9 & -      & 40.9                            \\
			BiSeNet (2018)         & 28.6  & 31.3  & 42.5 & -      & \textbf{12.8}                            \\ \midrule
			AWMF-CNN$^{\dag}$ (2019)        & 26.8  & 32.1  & 42.2 & 42.7   & 76.3                            \\
			AWMF-CNN$^{\ddag}$ (2019)        & 19.4  & 35.5  & 37.1 & 37.7   & 61.2                            \\ 
			AWMF-CNN$^{\dag}$ (fixed) (2019)        & 28.6  & 31.3  & 42.5 & 42.9   & 76.3                            \\
			AWMF-CNN$^{\ddag}$ (fixed) (2019)        & 25.2  & 31.7  & 37.9 & 38.7   & 61.2                            \\ \midrule
			MSN$^{\dag}$                  & 46.4  & 47.2  & 42.5 & 47.6   & 15.6                            \\
			MSN$^{\ddag}$                    & \textbf{37.8}  & \textbf{38.6}  & 37.9 & \textbf{39.1}   & 9.2                             \\
			MSN$^{\dag^{*}}$                  & \textbf{47.2}  & \textbf{47.9}  & 42.5 & \textbf{48.1}   & 15.6                            \\ \bottomrule
	\end{tabular}}
	\caption{Comparison results on BACH. $X_1$ and $X_2$ are the non-meta-branches in our method, and $X_3$ is the trained meta-branch.}
	\label{com_res_bach}
\end{table}
\subsection{Comparisons with State-of-the-art Methods}\label{com_res}
\paragraph{Result on BACH Dataset.} We compare our method with five state-of-the-art methods: UNet \cite{UNet}, PSPNet \cite{PSPNet}, BiSeNet \cite{BiSeNet}, DeepLab-V3+ \cite{DeepLab-V3+} and AWMF-CNN \cite{AWMF-CNN}, where the first four methods are representative general semantic segmentation frameworks, and the last one is the latest powerful multi-branch method for processing medical URIs. Because the first four methods are not the multi-branch structure, the fusion results are not available, which are denoted by ``-" in Table \ref{com_res_bach}. All methods have publicly provided code except AWMF-CNN, thus we reproduce it using Pytorch. Moreover, in \cite{AWMF-CNN}, AWMF-CNN uses UNet as backbone, we also implement our method using the same backbone without loss of generality. We train all competitors by using Eq. (\ref{loss function}) on the training set. For the first four methods we train each model with a specific resolution for $30$ epochs. For AWMF-CNN, we adopt two training ways: 1) Similar to the original way in AWMF-CNN, firstly we pretrain its three branches for $10$ epochs, then train the fusion parts. After that, we alternately train the multi-resolution branches and fusion part for 20 epochs. 2) we only train its fusion part for $30$ epochs with the fixed trained branches, we denote it as AWMF-CNN (fixed). For the convenience of expression, we use different marks in the superscript to denote the different settings: ``$^{\dag}$'': use BiSeNet as backbone; ``$^{\ddag}$'': use UNet as backbone; ``$^{*}$'': similar to other comparison methods, we train MSN on the training set. For all methods, we report the best results in the test set. 

As shown in Table \ref{com_res_bach}, we observe that our method achieves the best results. Note that, with the help of our special weight sharing mechanism, we improve the result significantly for the non-meta-branches by almost $10\%$ mIoU compared with the counterpart branches of BiSeNet and UNet, respectively (For example, the branch of $X_1$ of MSN$^{\dag}$ achieves $46.4\%$ mIoU, while the one of BiSeNet only get $28.6\%$). Meanwhile, the result can be further boosted with our meta-fusion.   

It can be also found that our method already obtained the best results by only training on the small sub-training set, {\it e.g.,} MSN$^{\dag}$, while other comparison methods are trained on the training set. When we also train on training set, {\it e.g.,} MSN$^{\dag^*}$, we can get better performance. Therefore, it can be concluded that MSN is more flexible in data requirements.

Another important point is that, the amount of parameters of MSN is almost the same as that of a single network (see MSN$^{\dag}$ vs BiSeNet and MSN$^{\ddag}$ vs UNet), and is much smaller than AWMF-CNN, thus our model has lower complexity and is more practical.
\begin{table}[!t]
	\centering
	\resizebox{.95\columnwidth}!{\begin{tabular}{@{}lrrrrr@{}}
			\toprule
			\multirow{2}{*}{Method} & \multicolumn{4}{c}{mIoU (\%)} & \multirow{2}{*}{$\#$ Parm (M)} \\ \cmidrule(lr){2-5}
			& $X_1$    & $X_2$    & $X_3$   & Fusion &                                 \\ \midrule
			DeepLab-V3+ (2018)      & 42.8  & 47.7  & \textbf{48.1} & -      & 40.9                            \\
			BiSeNet (2018)         & 45.1  & 46.4  & 46.1 & -      & \textbf{12.8}                            \\ \midrule
			AWMF-CNN$^{\dag}$ (2019)        & 43.8  & 42.7  & 45.9 & 45.5   & 76.3                            \\
			AWMF-CNN$^{\dag}$ (fixed) (2019)        & 45.1  & 46.4  & 46.1 & 48.9   & 76.3                            \\
			\midrule
			MSN$^{\dag}$                  & 49.0  & 48.8  & 46.1 & 49.4   & 15.6                            \\
			MSN$^{\dag^{*}}$                  & \textbf{52.8}  & \textbf{48.5}  & 46.1 & \textbf{54.6}   & 15.6                            \\ \bottomrule
	\end{tabular}}
	\caption{Comparison results on ISIC.}
	\label{com_res_isic}
\end{table}
\paragraph{Result on ISIC Dataset.} The comparison results on ISIC are shown in Table \ref{com_res_isic}. For fast implementation without loss of generality, we compare MSN with the latest three methods: DeepLab-V3+, BiSeNet and AWMF-CNN. MSN also achieves the best result with the comparable amount of parameters compared to other methods.

\begin{figure}[!t]
	\center{\includegraphics[width=.95\columnwidth]{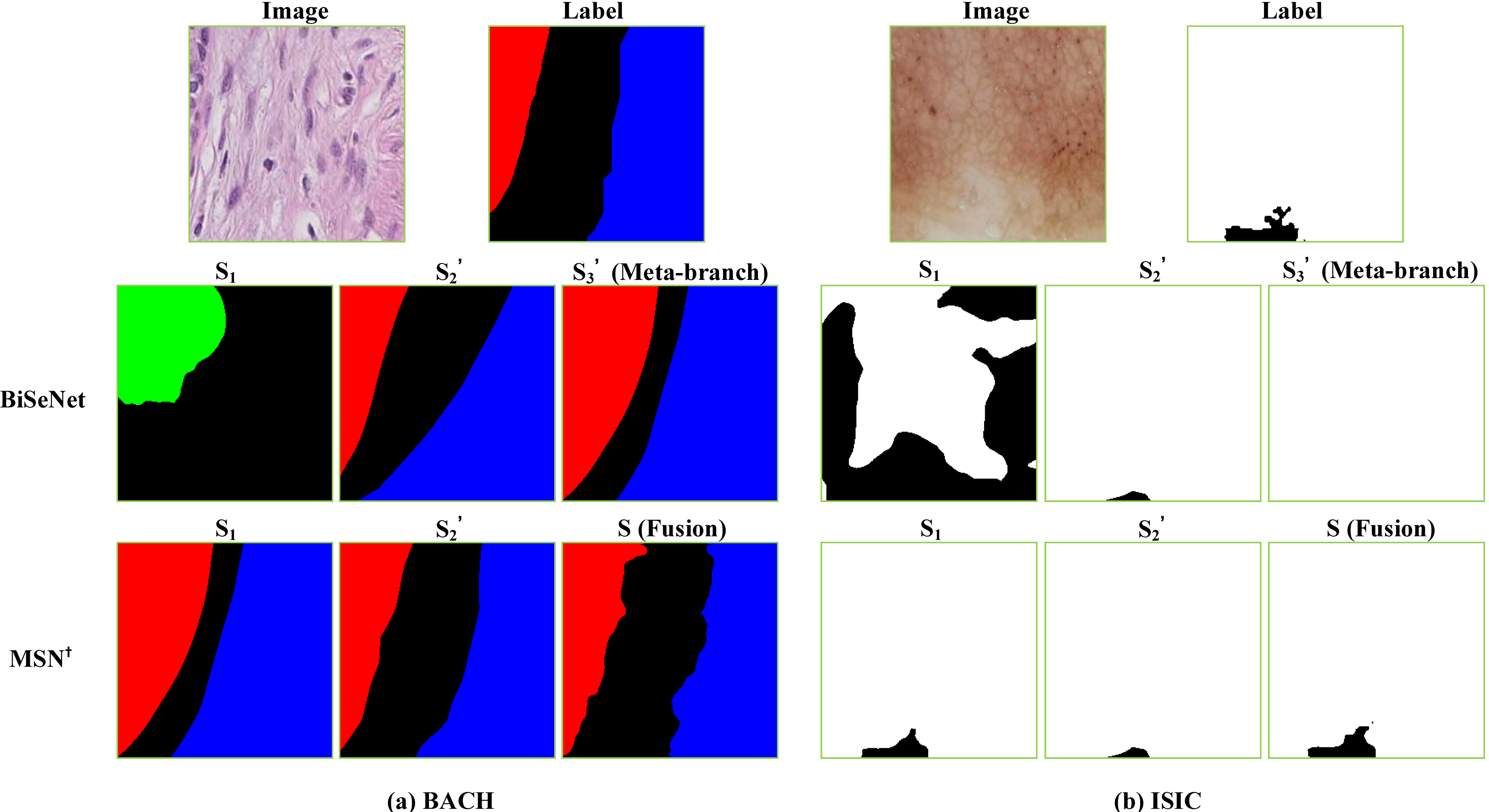}}
	\caption{The visualization results on BACH and ISIC. The first row contains the examples of image patches and counterpart labels. The second row is the results of our backbone BiSeNet, where three branches are trained separately. The third row is ours, where the first two columns of each dataset are our non-meta-branches, and the last column is our fusion result.}
	\label{visual_BACH_ISIC}
\end{figure}

\paragraph{Visualization.} Finally, we visualize the results on BACH and ISIC. Due to space limitation, we directly compare our method (BiSeNet as backbone) with BiSeNet that trains three branches separately. The results are illustrated in Fig. \ref{visual_BACH_ISIC}. Obviously, with the special weight sharing mechanism, the non-meta-branches of MSN significantly outperform all the branches of BiSeNet. More importantly, with our meta-fusion mechanism, some details can be further refined, which makes the final result more complete.

\subsection{Ablation Study} \label{ablation-study}
\paragraph{Effectiveness of Weight Sharing Mechanism.} The special weight sharing mechanism can not only significantly reduce the amount of parameters of multi-branch model, but also realize the knowledge transfer between the branches. Thus the results of the non-meta-branches can be promoted on the basis of the meta-branch, which has been verified in Section \ref{com_res}. To further verify its effectiveness, we design the following experiments:

Firstly, we compare four methods: (1) Meta-branch: we only use the trained meta-branch to obtain the results of all resolution inputs, without fixing the gap layers. (2) Multi-branch: all branches in this structure are separately trained from scratch. 
(3) MSN$^{\dag}$ and (4) MSN$^{\dag^*}$. The backbone of all methods are BiSeNet. For fair comparison, we also conduct our meta-fusion mechanism on the first two compared methods. The results are shown in Table \ref{abltion_study_eff_weight_sharing}. It is observed that our non-meta-branches outperform other methods significantly, it shows that our weight sharing mechanism can effectively eliminate the gaps between meta-branch and other branches, then improve the performance by leveraging existing knowledge. And the fusion results are based on branches results, therefore, the improvement of branches is also conducive to the improvement of final performance.
\begin{table}[!t]
	
	\centering
	\begin{tabular}{@{}lrrrrr}
		\toprule
		\multirow{2}{*}{Method} & \multicolumn{4}{c}{mIoU (\%)}  \\ \cmidrule(lr){2-5}
		& $X_1$    & $X_2$    & $X_3$   & Fusion &                                 \\ \midrule
		Meta-branch      & 21.1  & 31.8   & 42.5 & 39.3                               \\
		Multi-branch        & 28.6  & 31.3  & 42.5 &  44.4                                \\ 
		
		MSN$^{\dag}$                  & 46.4  & 47.2  & 42.5 & 47.6                            \\
		MSN$^{\dag^{*}}$                  & \textbf{47.2}  & \textbf{47.9}  & 42.5 & \textbf{48.1}                           \\ \bottomrule
	\end{tabular}
	\caption{The effectiveness of the weight sharing of MSN on BACH.}
	\label{abltion_study_eff_weight_sharing}
\end{table}

\begin{figure}[!t]
	\center{\includegraphics[width=.95\columnwidth]{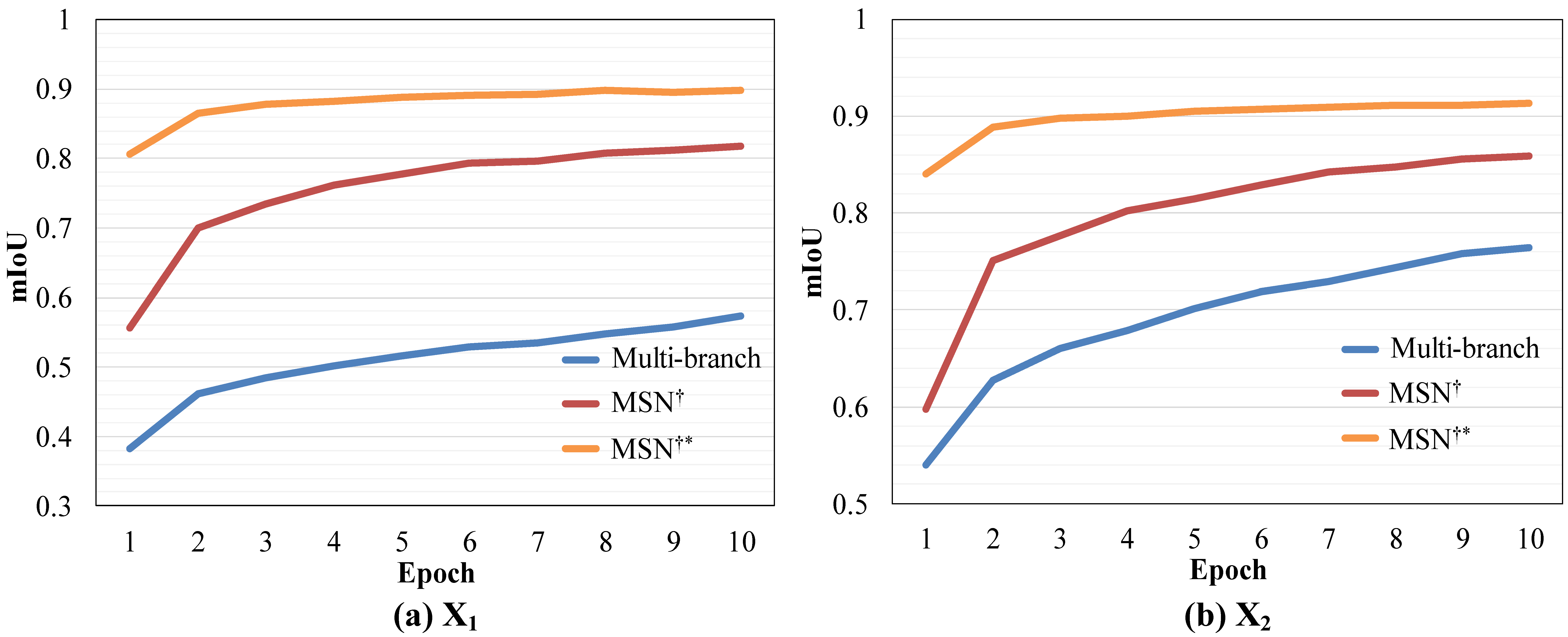}}
	\caption{The train trend on BACH of non-meta-branches, {\it i.e.,} the branches of $X_1$ and $X_2$.}
	\label{convergence_x1x2}
\end{figure}

Secondly, we compare the convergency of the non-meta-branches of MSN and the ones of Multi-branch (training on training set). The curves are illustrated in Fig. \ref{convergence_x1x2}. It shows that our method not only performs on segmentation better, but converges faster. And when we use the training set rather than the sub-training set for training (MSN$^{\dag^*}$), we can obtain better convergence performance.

\begin{table}[!t]
	\centering
	\begin{tabular}{@{}lrrrrr}
		\toprule
		\multirow{2}{*}{Method} & \multicolumn{4}{c}{mIoU (\%)}  \\ \cmidrule(lr){2-5}
		& $X_1$    & $X_2$    & $X_3$   & Fusion &                                 \\ \midrule
		on non-gap layers      & 33.7  & 30.3  & 42.5 & 34.5                              \\
		
		MSN$^{\dag}$                  & \textbf{46.4}  & \textbf{47.2}  & 42.5 & \textbf{47.6}                            \\
		\bottomrule
	\end{tabular}
	\caption{The impact of gap layers on our weight sharing. `on non-gap layers' denotes that we add Mem-FP and Mem-RM only to the non-gap layers.}
	\label{abltion_study_eff_weight_sharing}
\end{table}

Thirdly, to explore the impact of `gap layers', we attempt to only add Mem-RM at the `non-gap layers', {\it i.e.,} the layers whose index in black font in Fig. \ref{mean_variance_bise}. As expected, the result of this approach drops a lot, which shows that our effort to fix the gaps between meta-branch and other branches is reasonable.
\begin{table}[!t]
	\centering
	\resizebox{.95\columnwidth}!{\begin{tabular}{@{}llrrrrr@{}}
		\toprule
		\multirow{2}{*}{Dataset} & \multirow{2}{*}{Method} & \multicolumn{4}{c}{mIoU (\%)}                                                                                                  & \multirow{2}{*}{\# F (M)}         \\ \cmidrule(lr){3-6}
		&                         & X1                    & X2                       & \multicolumn{1}{r}{X3}                     & Fusion                         &                                  \\ \midrule
		\multirow{3}{*}{BACH}    & w/o Meta                & \multirow{3}{*}{46.4} & \multirow{3}{*}{47.2}    & \multicolumn{1}{l|}{\multirow{3}{*}{42.5}} & 23.7                           & \textbf{0.0004} \\
		& AWMF-CNN                &                       &                          & \multicolumn{1}{c|}{}                      & 47.1                           & 37.9                             \\
		& MSN                     &                       &                          & \multicolumn{1}{c|}{}                      & \textbf{47.6} & 0.01                             \\ \midrule
		\multirow{3}{*}{ISIC}                     & w/o Meta                & \multirow{3}{*}{49.0}                  & \multirow{3}{*}{48.8} & \multicolumn{1}{c|}{\multirow{3}{*}{46.1}}                  & 36.8       & \textbf{0.0001}       \\
		& AWMF-CNN                &                       & \multicolumn{1}{l}{}     & \multicolumn{1}{c|}{}                      & 48.2       & 37.9         \\
		& MSN                     &                       & \multicolumn{1}{l}{}     & \multicolumn{1}{c|}{}                      & \textbf{49.4}       & 0.006       \\ \bottomrule
	\end{tabular}}
	\caption{The effectiveness of meta-fusion on BACH and ISIC. \# F denotes the parameter amount of the fusion part of each method.}
	\label{meta_fusion_bach_isic}
\end{table}

\paragraph{Effectiveness of Meta-fusion.} To verify the effectiveness of the meta-fusion mechanism, we compare three methods: (1) w/o Meta: the same fusion structure as ours without meta-fusion mechanism, {\it i.e.,} two $3\times3$ convolution layers in Fig. \ref{MSN}, and we train it end-to-end from scratch.  (2) AWMF-CNN: the fusion mechanism in AWMF-CNN, which introduces a heavy weighting net for branches weighting, then uses some convolution layers training from scratch for fusion. (3) MSN: our meta-fusion mechanism. For fair comparison, we fix the results of three resolutions, which come from our trained three branches whose backbones are BiSeNet, and then we train the fusion part of all comparison methods on the sub-training set. 

The comparison results on BACH and ISIC are shown in Table \ref{meta_fusion_bach_isic}. We can observe that our method outperform w/o Meta's significantly. Although we have more parameters than it, it is almost negligible due to the small order of magnitude. The performance of AWMF-CNN's is a little lower than ours, but its fusion structure is more complicated than ours, resulting in a sharp increase in the amount of parameters.

\begin{figure}[!t]
	\center{\includegraphics[width=.95\columnwidth]{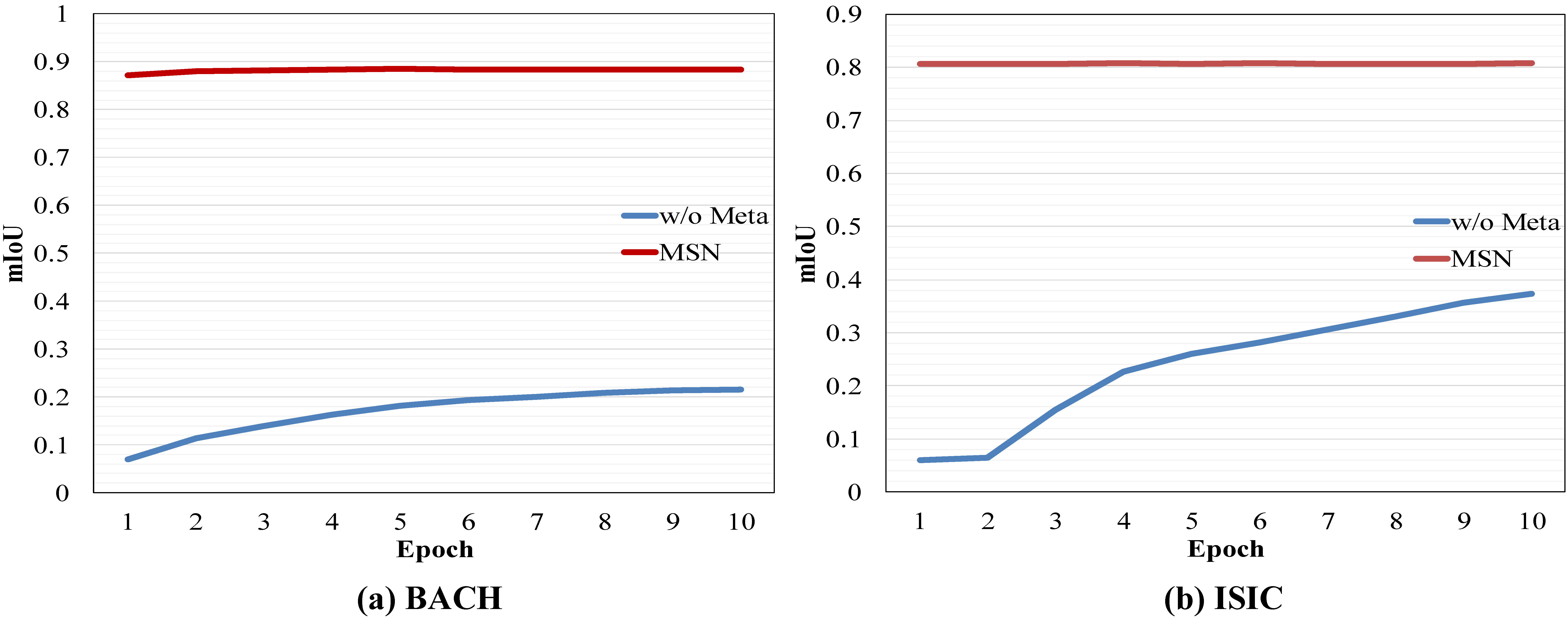}}
	\caption{The train trend of our meta-fusion and the fusion of Non-meta on the sub-training set of BACH and ISIC.}
	\label{meta-fusion_trend}
\end{figure}

We further illustrate the train trend of meta-fusiion mechanism and w/o Meta's on BACH and ISIC in Fig. \ref{meta-fusion_trend}. As expected, our method achieves an extremely better convergency, {\it e.g.,} it converges in almost $1$ epoch.
\section{Conclusions}
In this work, we propose MSN for the effective segmentation of medical URIs. A novel meta-fusion module with a very simple but effective structure is introduced for branches fusion through a meta-learning way. Moreover, MSN achieves a lightweight multi-branch structure with the help of our particular weight sharing mechanism. The experimental results on BACH and ISIC demonstrate that our method achieves the best comprehensive performance.

\bibliographystyle{named}
\bibliography{ijcai20}

\end{document}